\documentclass[runningheads]{llncs}

\usepackage{eccv}

\usepackage{eccvabbrv}

\usepackage{graphicx}
\usepackage{booktabs}

\newcommand{\vh}{{\bf h}}

\newcommand{\vp}{{\bf p}}

\newcommand{\vE}{{\bf E}}

\newcommand{\vI}{{\bf I}}
\newcommand{\vM}{{\bf M}}

\newcommand{\vP}{{\bf P}}

\newcommand{\vW}{{\bf W}}
\newcommand{\vV}{{\bf V}}
\newcommand{\vS}{{\bf S}}
\newcommand{\vT}{{\bf T}}
\newcommand{\vX}{{\bf X}}

\def\onedot{. }
 
\def\ie{\emph{i.e}\onedot}

\usepackage{color}
\usepackage{colortbl}
\usepackage{multirow, makecell}
\usepackage{array}
\usepackage{wrapfig}
\usepackage{cutwin}
\usepackage{marvosym}
\newcolumntype{C}[1]{>{\centering\arraybackslash}p{#1}}
\usepackage[accsupp]{axessibility}  

\usepackage[pagebackref,breaklinks,colorlinks,citecolor=eccvblue]{hyperref}

\usepackage{orcidlink}

\begin{document}

\title{Cross-Platform Video Person ReID: A New Benchmark Dataset and Adaptation Approach} 

\titlerunning{Cross-Platform Video Person ReID}

\author{
Shizhou Zhang\inst{1}\orcidlink{0000-0002-5914-7109} \and
Wenlong Luo\inst{1}\orcidlink{0009-0000-5439-2749} \and
De Cheng\inst{2}\thanks{Corresponding author, dcheng@xidian.edu.cn}\orcidlink{0000-0003-1932-4390}\and
Qingchun Yang\inst{1}\orcidlink{0009-0005-2822-2443}\and
Lingyan Ran\inst{1}\orcidlink{0000-0002-3084-9860}\and
Yinghui Xing\inst{1}\orcidlink{0000-0001-6021-8261}\and
Yanning Zhang\inst{1}\orcidlink{0000-0002-2977-8057}
}

\authorrunning{S.Zhang et al.}

\institute{Northwestern Polytechnical University, Xi'an Shaanxi 710000, China  \and
Xidian University, Xi'an Shaanxi 710000, China
\email{\{szzhang,lran,xyh\_7491,ynzhang\}@nwpu.edu.cn}
\email{\{luowenlong,yqc123\}@mail.nwpu.edu.cn}
\email{\{dcheng\}@xidian.edu.cn}}

\maketitle

\begin{abstract}
  In this paper, we construct a large-scale benchmark dataset for Ground-to-Aerial Video-based person Re-Identification, named G2A-VReID, which comprises 185,907 images and 5,576 tracklets, featuring 2,788 distinct identities.
  To our knowledge, this is the first dataset for video ReID under Ground-to-Aerial scenarios. G2A-VReID dataset has the following characteristics: 1) Drastic view changes; 2) Large number of annotated identities; 3) Rich outdoor scenarios; 4) Huge difference in resolution. Additionally, we propose a new benchmark approach for cross-platform ReID by transforming the cross-platform visual alignment problem into visual-semantic alignment through vision-language model ($i.e.$, CLIP) and applying a parameter-efficient Video Set-Level-Adapter module to adapt image-based foundation model to video ReID tasks, termed VSLA-CLIP. Besides, to further reduce the great discrepancy across the platforms, we also devise the platform-bridge prompts for efficient visual feature alignment. Extensive experiments demonstrate the superiority of the proposed method on all existing video ReID datasets and our proposed G2A-VReID dataset. The code and datasets are available at \href{https://github.com/FHR-L/VSLA-CLIP}{https://github.com/FHR-L/VSLA-CLIP}.
  
\end{abstract}

\section{Introduction}
Video-based person Re-Identification (VReID)~\cite{STA, M3D, GLTR, SINet}, has been attracting much attention in recent years, as video can provide richer information than single image.
Existing research efforts on video-based ReID are mostly based on data from the same platforms, such as ground surveillance cameras.
Suppose that a suspect has committed a crime in the city where abundant surveillance cameras have been deployed and escaped into the rural areas where there are no deployed ground surveillance cameras in advance.
One feasible solution is sending a moving camera with the help of an airbone UAV platform.
Thus, the technical crux has been turned into the cross-platform video-based person ReID.

In this paper, to meet the research need of cross-platform video person ReID, we construct a large-scale benchmark dataset named Ground-to-Aerial Video ReID (G2A-VReID). 
The G2A-VReID dataset consists of 185,907 images in total, with 5,576 tracklets belonging to 2,788 different person IDs. 
Each person ID includes two tracklets captured by the UAV and ground surveillance platforms, respectively. 
There is an average of 33.3 images for each tracklet. The scale of G2A-VReID dataset is larger than most existing video-based person ReID datasets such as MARS~\cite{mars}, iLIDS~\cite{iLIDS}, PRID-2011~\cite{PRID}, etc.

To capture the videos of the same person by both the ground surveillance camera and the UAV-mounted camera, we simulate the ground-to-aerial platform ReID by fixating a ground surveillance camera at a specific location, while flying a DJI consumer UAV nearby.
The ground camera is set at about 2.0 meters above the ground, and the flight altitudes of UAVs vary from 20 meters to 60 meters. Additionally, to be more realistic, the flight mode is adjusted randomly among hovering, cruising, and rotating with diverse view angles which greatly enriches the perspectives of the dataset.

Furthermore, the dataset is collected at nine different scenarios, including school campuses, subway station entrances, tourist sites, crossroads, etc.
As shown in Fig.~\ref{dataset}, the cross-platform video person ReID task is much more challenging than the counterpart in single ground platform, as the tracklets captured in the ground to aerial cross-platform scenarios are featured in drastic variations of view-points, poses, and resolutions.
We have evaluated nine existing video-based person ReID algorithms on our newly collected cross-platform dataset. The experimental results showed inferior performances compared with those conventional single-platform datasets. Due to the great challenges of drastic view, pose, and resolution changes, it is not easy to align the visual part features between the cross-platform devices, which is essential in ReID task.

Recently, with the emergence of large-scale pre-trained vision-language models, $e.g.$, CLIP~\cite{CLIP}, a well-aligned visual-semantic space can be obtained through cross-modality contrastive learning of large web visual data along with high-level language descriptions.
Although for the ReID task, there is no language descriptions for each person whose identity is just denoted as an index number, a set of learnable description tokens can also be introduced to roughly describe each ID~\cite{CLIP_ReID}.
In this paper, we propose to transform the cross-platform visual alignment problem into visual-semantic alignment with the help of the foundation model CLIP.
To be concrete, a two-stage optimization strategy is utilized, which aims to learn description tokens for each ID in the first stage, and fine-tunes the Image Encoder with aligning visual embeddings to semantic features obtained through the learned description token in the second stage.
Our experiments demonstrate that fine-tuning the Image Encoder with the constraint of visual-semantic alignment achieves competitive performance. 

However, there are two obvious drawbacks in adapting image-based pre-trained foundation models to video ReID tasks by simply fine-tuning. One is the huge training cost with large-scale trainable parameters, and another is that the image encoder lacks the capability of modeling inter-frame information.
Many previous works\cite{3DPeS, STRF, BiCnet_TKS} deem video as a stack of frames with temporal structure, and are devoted to modeling temporal features with well-designed modules. 
But these works ignore the complementarity of frames in a video, which proved to be more effective in ReID task\cite{SINet}. 
Moreover, from the aerial perspective, temporal information is limited due to severe self-obstruction. As shown in Tab.~\ref{tab:compare_with_SOTA}, temporal models\cite{AP3D, STMN, BiCnet_TKS} show inferior performance on G2A-VReID.
In this paper, we present a new perspective that regards a video clip as a disordered set and propose a parameter-efficient Video Set-Level-Adapter (VSLA) module for foundation modal adaptation. Concretely, VSLA consists of a Cross-Frame Attention Adapter (CFAA) and an Intra-Frame Adapter (IFA). 
CFAA uses cross-frame attention to allow information exchange between frames, enabling our model to collect complementary features in each video set for powerful video-level representations.
IFA transfers the visual ability of image-based foundation model to downstream tasks, providing strong intra-frame appearance representation.

Furthermore, we also propose the Platform Bridging Prompt (PBP) module to solve the visual misalignment problem in cross-platform tasks, where the prompts are adopted to provide explicit instruction to the pre-trained models for generating task-specific results\cite{prompt1, JiaTCCBHL22, li2023boosting, 10171397}. Specifically, the designed PBP is two sets of platform-specific prompts brought in Image Encoder, which aims to guide the model to focus on learning platform-invariant features, thus bridging the semantic gap of visual features between the ground and aerial platforms.  

In summary, the main contributions are as follows:
\begin{itemize}
    \item We are the first to collect a large-scale Ground-to-Aerial Video person ReID benchmark dataset for the task of cross-platform video-based person ReID and conducted extensive baseline methods on our dataset.
    \item We propose to transform the essential cross-platform visual part alignment problem into visual-semantic alignment with the help of CLIP, and propose PBP to further bridge the semantic gap of visual features between the ground and aerial platforms. 
    \item We propose the Video Set-Level-Adapter to efficiently adapt pre-trained image-based visual foundation model to the video ReID tasks. Our methods achieves state-of-the-art performances on three widely used video ReID datasets and our cross-platform benchmark dataset.
\end{itemize}

\section{Related Works}
In this section, we provide a concise review of two sets of works closely related to our research.

\textbf{Video ReID Datasets.}
Existing works on person ReID can be categorized into image-based ReID\cite{9019850, cheng2024continual, CLIP_ReID, 9585033, cheng2023efficient, sie_transreid, HDCP, zhu2023aaformer} and video-based ReID\cite{M3D, STRF, Snippet, mars}. 
For video-based ReID, the popular datasets include PRID-2011\cite{PRID}, iLIDS\cite{iLIDS}, MARS\cite{mars} and LS-VID\cite{GLTR_LS-VID}, etc. 
PRID-2011 comprises multiple person trajectories captured by two static surveillance cameras, encompassing only 400 sequences involving 200 individuals.
In contrast, LS-VID is a large-scale benchmark featuring 14,943 sequences of 3,772 persons, with videos captured at various times throughout the day. 
Many works have achieved superior performances on these datasets. 
Specifically, FGReID\cite{FG-ReID} achieved Rank-1 at 96.1\% on PRID-2011, SINet\cite{SINet} got 92.5\% of Rank-1 on iLIDS and DenseIL\cite{DenseIL} achieved an mAP of 87.0\% on MARS, indicating a saturation trend on these datasets.
The existing datasets are all captured with a single platform, i.e. ground surveillance cameras, while we aim to collect a Ground-to-Aerial cross-platform video ReID dataset to support the development of this field.

\textbf{Video ReID Methods.} The object processed in video-based person ReID is a video composed of a sequence of person images. Videos contain richer temporal and spatial information than images. Previous works used 3D CNNs\cite{M3D, STRF, AP3D}, temporal weighting\cite{STA, SeeForest, Snippet, AP3D, mars}, optical flow\cite{feichtenhofer2016convolutional, liu2017video, chung2017two} and many other methods\cite{BiCnet_TKS, STMN, TCLNet, AFA} to model the spatiotemporal information of video sequences to alleviate the negative effects of appearance change, occlusion, pose variation, etc.
For 3D CNNs, STRF\cite{STRF} proposed a trainable unit with negligible computational overhead, which is used in conjunction with 3D-CNN to learn discriminate 3D features.
For temporal weighting, 
AP3D\cite{AP3D} assigns attention scores for each spatial region to achieve discriminative parts mining and frame selection. 
Optical flow refers to the movement of target pixels in an image due to the movement of objects in the image or the movement of the camera in two consecutive frames. 
STA\cite{STA} makes use of color and optical flow information in order to capture appearance and motion information. 
An essential topic to improve the performance of video-based ReID is the visual part alignment between query and gallery videos. PiT\cite{PiT} divides each frame into small patches of different granularity in different directions, allowing the model to align two videos with multi-scale local information.
It is relatively easy to align the visual part features between the query and gallery videos for these methods by utilizing a simple stripe partition, as the variations of view, pose, and resolution are limited among the single ground cameras.

To solve the severe misalignment of visual features in cross-platform tasks, we resort to visual-semantic alignment of the CLIP model to align the cross-platform person features.

\section{Dataset}

In this section, we first introduce how we collect and annotate our G2A-VReID dataset in Sec.~\ref{dataset_collect} and Sec.~\ref{annotation}. Then, we make comparisons with other datasets and highlight the key characteristics of G2A-VReID in Sec.~\ref{dataset_comparison}.
\begin{figure}[t]
\begin{minipage}[t]{0.4\linewidth}
    \centering
    \includegraphics[width=\textwidth]{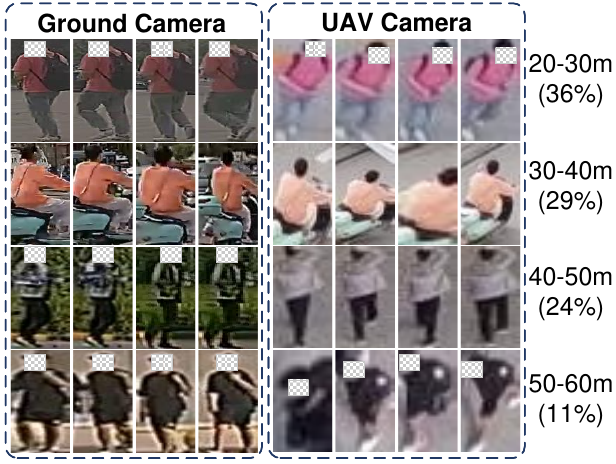}
	\caption{Visualization of proposed G2A-VReID at different heights.}
	\label{dataset}
\end{minipage}
\begin{minipage}[t]{0.6\linewidth}
    \centering
	\includegraphics[width=0.8\linewidth]{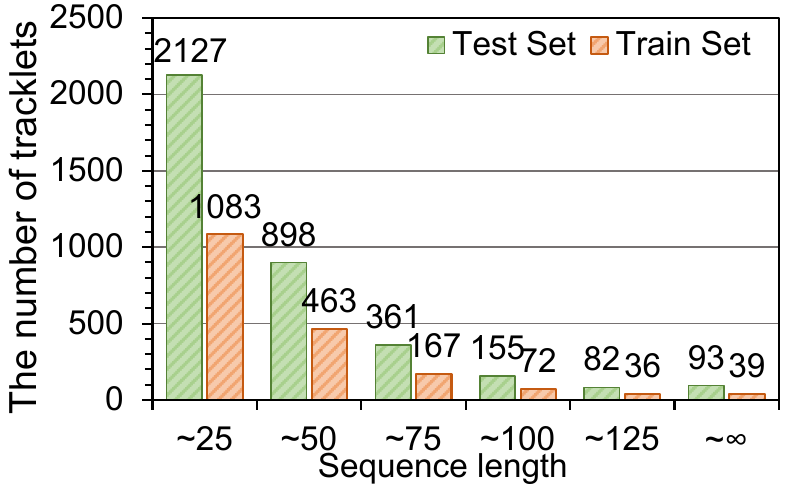}
	\caption{The distributions of sequence length.}
	\label{dataset_stat}
\end{minipage}
\end{figure}

\subsection{Dataset Collection}
\label{dataset_collect}

To increase the richness of data and make it closer to the real environment. The videos are captured from 9 different scenarios, including library, crossroads, bus stop, tourist sites, etc. Ground surveillance cameras are used to shoot videos from the ground perspective, and a DJI Mavic UAV is adopted to gather videos from the sky perspective. In detail, the surveillance camera is fixed at a height of about two meters above the ground, and the UAV flies at different heights from 20 to 60 meters. The UAV flies in a mode of hovering, cruising, and rotating, making the captured persons contain richer perspectives. 
We cropped the captured video at intervals of 0.5 seconds to generate 31,770 frames. 
As shown in Fig.~\ref{dataset}, there are great differences in the viewing perspective and resolution of images taken on different platforms, making it more challenging than existing datasets.

\subsection{Annotation}
\label{annotation}
During annotation, all persons appeared in the videos are marked with boundary boxes, and each person is cropped from the scene image according to the box. At the same time, we use mosaic to mask the clear face information for privacy protection.
Then, the same people in the UAV and surveillance videos are associated and assigned unique IDs.
Next, we combine all the images of a person in one camera into one trajectory. Thus, each person has at least two trajectories, one from the surveillance camera and the other from the UAV.
Finally, we annotated 185,907 images of 2,788 identities, corresponding to 5,576 tracklets. 
Fig.~\ref{dataset_stat} shows the distributions of sequence length.

\subsection{Characteristics of Our G2A-VReID}
\label{dataset_comparison}
Compared with existing VReID datasets~\cite{GLTR_LS-VID, mars, iLIDS} , the characteristics of G2A-VReID are as follows:
\textbf{1) Drastic view changes.} 
The tracklets in the query and gallery sets are captured from different types of cameras. Consequently, the transitions between the views in the query and gallery tracklets are significantly different.
\textbf{2) Large number of annotated identities.} Our G2A-VReID consists of 2,788 person IDs and 185,907 images, corresponding to 5,576 tracklets. The number of identities is significantly higher than all existing datasets except LS-VID~\cite{GLTR_LS-VID}, as shown in Tab.~\ref{tab:dataset_comparison}.
\textbf{3) Rich outdoor scenarios with large view changes.} The G2A-VReID consists of footage from nine diverse scenarios. This diversity enables G2A-VReID to accurately represent realistic environments for person ReID. In contrast, the videos from Mars~\cite{mars} are captured on a university campus, while iLIDS~\cite{iLIDS} only contains videos collected from an airport arrival hall. 
\textbf{4) Huge difference in resolution.} 
As depicted in Fig.~\ref{dataset}, the height of the UAV-mounted camera varies significantly, spanning from 20 to 60 meters.
The width distribution of individuals in images captured by ground cameras primarily ranges from 10 to 70 pixels. Whereas, in UAV-captured images, this range is narrower from 5 to 35 pixels.

\begin{table*}[t]
\normalsize
\centering
\scriptsize
\caption{\footnotesize Comparison of G2A-VReID with other Video-ReID datasets. \textbf{CWM} denotes the camera working mode. \textbf{AD} is the average duration of each video sequence.}
\begin{tabular}[t]{C{1.6cm}|C{1.6cm}C{1.5cm}C{1.5cm}C{1.5cm}C{1.9cm}C{1.2cm}}
\toprule
Datasets & G2A-VReID & LS-VID\cite{GLTR_LS-VID} & Mars\cite{mars} & iLIDS\cite{iLIDS} & PRID-2011\cite{PRID} & 3DPeS\cite{3DPeS} \\ \midrule
identities & 2,788 & 3,772 & 1,261 & 300 & 200 & 200  \\ 
tracklets & 5,576 & 14,943 & 20,715 & 600 & 400 & 1,000   \\ 
images & 185,907 & 2,982,685 & 1,067,516 & 42,460 & 40,033 & 200,000  \\ 
AD (s) & \textbf{16.7} & 6.7 & 5.6 & 2.4 & 3.3 & 6.7  \\ 
camera & 2 & 15 & 6 & 2 & 2 & 8   \\ 
view & \textbf{ground$\mathbf{~\&~}$sky} & ground & ground & ground & ground & ground  \\ 
CWM & \textbf{moving} & fixed & fixed & fixed & fixed & fixed   \\ \midrule
\end{tabular}
\label{tab:dataset_comparison}
\end{table*}

\subsection{Privacy Protection}
We try our best to protect the privacy of pedestrians from the following aspects:
1) We mask the faces of all pedestrians using a mosaic to eliminate privacy information, effectively minimizing privacy risks.  
2) We use cordons to mark data collection areas and post notifications near the sites during the data capture process. 
However, we admit the limitation that we can not ensure every pedestrian is informed.
3) The dataset will be licensed for non-profit academic research only. 
More details about privacy protection (\textit{e.g.} notification, mosaic, and license) are available at \href{https://github.com/FHR-L/G2A-VReID}{https://github.com/FHR-L/G2A-VReID}.





\section{Approach}
\label{sec:approch}
\begin{figure*}[t]
	\centering
	\includegraphics[width=0.9\textwidth]{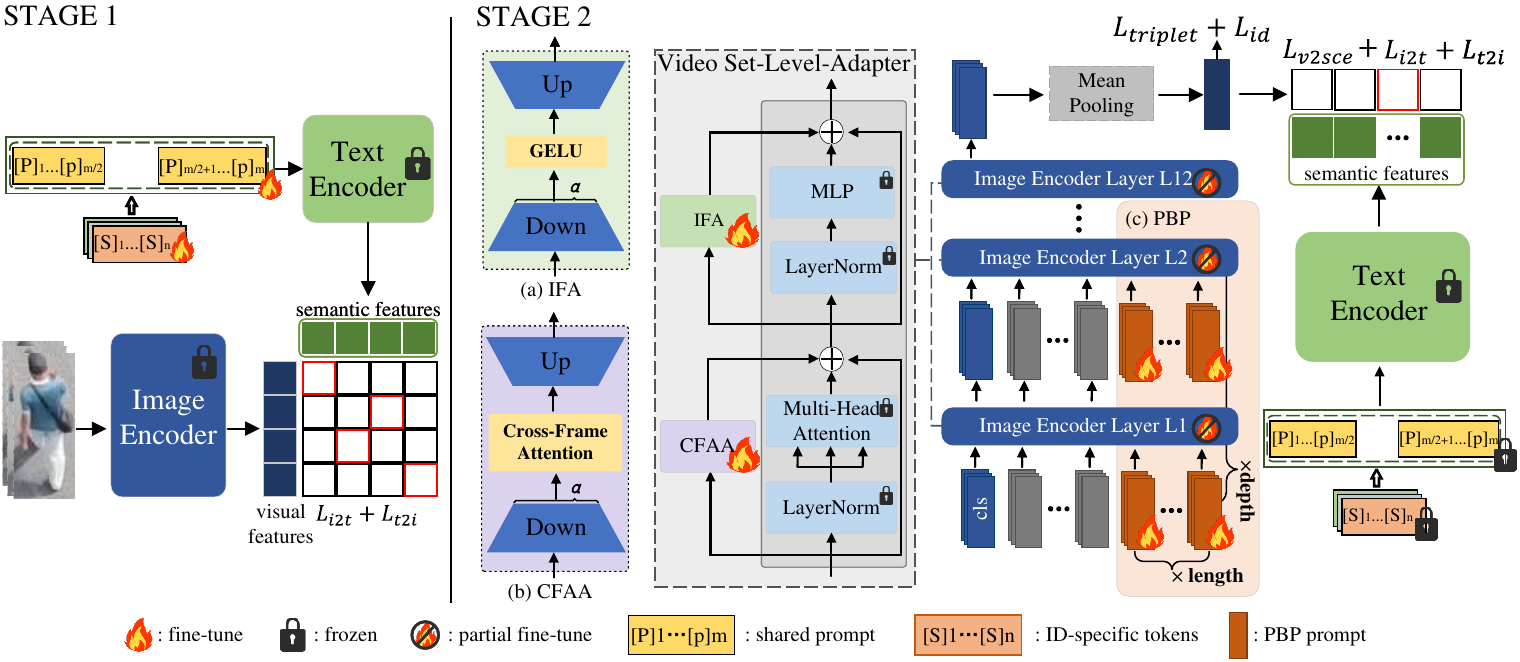}
	\caption{Overview of our proposed framework. ID-specific descriptions and shared text prompts are learned in stage one (left). Video Set-Level-Adapter and PBP are introduced and trained in the second stage (right) while freezing other parameters.}
	\label{overall_model}
\end{figure*}

Fig.~\ref{overall_model} illustrates the overall architecture of our proposed method. 
Our approach focuses on cross-platform video person ReID and aims to parameter-efficiently adapt pre-trained image-based visual foundation models to video person ReID tasks.
To bridge visual misalignment in cross-platform tasks, we propose to transform the fundamental visual alignment problem into visual-semantic alignment based on CLIP. Specifically, we design a simple baseline method, named FT-CLIP, through fine-tuning the Image Encoder of CLIP.
A two-stage training strategy is employed to optimize our approach. ID-specific description tokens are learned from samples originating from various platforms in the first training stage.
Then in the second stage, visual features extracted from different platforms are aligned with the semantic features obtained through the learned description tokens.
Our work shows that FT-CLIP with the constraint of visual-semantic alignment yields competitive performance, but it is not parameter efficient and ignores inter-frame information. 
Therefore, we propose the Video Set-Level-Adapter for efficient model tuning, termed as VSLA-CLIP, which outperforms FT-CLIP while utilizing fewer parameters.
To further bridge the semantic gap in cross-platform tasks, we propose a prompt-based approach called Platform-Bridge Prompt (PBP).

\subsection{Revisiting CLIP-ReID}
CLIP-ReID~\cite{CLIP_ReID} is the pioneering approach that employs pre-trained vision-language models for image-based ReID. 
CLIP~\cite{CLIP} relies on text labels to generate text descriptions.
However, the labels in ReID tasks are indexes rather than specific text, which lacks the ability to depict detailed information about the corresponding persons. To solve this problem, CLIP-ReID uses a series of ID-specific learnable tokens to learn text descriptions and adapts a two-stage optimization strategy.

In the first training stage, only ID-specific tokens are optimized to learn text descriptions for each ID. Text $\mathcal{TD}$ that feeds into Text-Encoder $\vE_t(\cdot)$ is ``a photo of $\rm[\vX]_1...[\vX]_M$ person", where $[\vX]_i$ is the learnable tokens. Text embedding $\vT$ and image embedding $\vI$ are obtained by:
\begin{align}
\vT = \vE_t(\mathcal{TD}),~~\vI = \vE_i(\mathcal{I}),
\end{align}
where $\vE_i(\cdot)$ is the Image Encoder. 
The image-to-text contrastive loss $\mathcal{L}_{i2t}$ and text-to-image contrastive loss $\mathcal{L}_{t2i}$ are used to optimize $\rm[\vX]_1...[\vX]_M$. 
Since there are samples with the same ID in a batch, $\mathcal{L}_{t2i}$ in CLIP-ReID is defined as:
\begin{equation}\label{eq:lt2i_P}
\mathcal{L}_{t2i}({y_i}) = \frac{-1}{|P(y_i)|}\sum_{p\in P(y_i)} \log\frac{\exp(s(\vI_p,\vT_{y_i}))}{\sum_{a=1}^B \exp(s(\vI_a,\vT_{y_i}))},
\end{equation}
where $\vT_{y_i}$ represents the text embedding of ID-${y_i}$, $P(y_i)=\{p\in \{1...B\}, y_p = y_i\}$ is the set of positive samples for $\vT_{y_i}$ and $B$ represents the batch size. 
$\mathcal{L}_{i2t}$ is similar to $\mathcal{L}_{t2i}$.
The overall loss function of stage one $\mathcal{L}_{stage1}$ is as follows:
\begin{equation}\label{eq:stage1loss}
\mathcal{L}_{stage1} = \mathcal{L}_{i2t} + \mathcal{L}_{t2i}.
\end{equation}

In the second stage, the ID-specific tokens and Text-Encoder are frozen. Triplet loss $\mathcal{L}_{tri}$\cite{triplet_loss}, identity loss $\mathcal{L}_{id}$, and image-to-text cross-entropy loss $\mathcal{L}_{i2tce}$ are used to optimize CLIP Image Encoder. The $\mathcal{L}_{i2tce}$ is defined as follows:
\begin{equation}\label{eq:li2tce}
\mathcal{L}_{i2tce}(y) = \sum_{k=1}^N -q_k\log\frac{\exp(s(\vI_y,\vT_{y_k}))}{\sum_{{y_a}=1}^N \exp(s(\vI_y,\vT_{y_a}))},
\end{equation}
where $q_k$ denotes smooth label\cite{LS} in the target distribution of the $k_{th}$ ID, $s$ represents cosine similarity, and $N$ is the number of identities.

\subsection{Visual-Semantic Alignment}
We propose to transform the fundamental challenge of cross-platform visual alignment into visual-semantic alignment, and explore the efficacy of fine-tuning to adapt CLIP to video-based ReID tasks with visual-semantic alignment, named the model FT-CLIP.
As shown in Fig.~\ref{overall_model} (left), learnable ID-specific description tokens $\rm[\vS]_i$ and shared text prompts $\rm[\vP]_i$ are inserted into the Text-Encoder. All the tokens that feed into the Text-Encoder are concatenated as ``$[\rm[\vP]_1...[\vP]_{n/2}:\rm[\vS]_1...[\vS]_M:\rm[\vP]_{n/2+1}...[\vP]_{n}]$". 
Semantic features $\vT$ can be obtained by:
\begin{align}
{\vT} &= \vE_t([\rm[\vP]_1...[\vP]_{n/2}:\rm[\vS]_1...[\vS]_M:\rm[\vP]_{n/2+1}...[\vP]_{n}]),
\end{align}
where $[\cdot : \cdot]$ represents the concatenating operation, the dimensions of $\rm[\vP]_i$ and $\rm[\vS]_i$ are the same as that of the word embedding. 

Inspired by CLIP-ReID~\cite{CLIP_ReID}, we adopt a two-stage optimization strategy.
In the first optimization stage, we freeze both the Image Encoder and Text Encoder, using loss function $\mathcal{L}_{stage1}$ in Eq. \eqref{eq:stage1loss} to optimize ID-specific description tokens and the shared text prompts. 
In the second optimization stage, Image Encoder is trained to align the video embeddings to semantic features. Given a video sample $\mathcal{V}_i \in \mathbb{R}^{T \times H \times W \times 3}$ with $T$ frames, the CLIP image encoder encodes the $T$ frames independently and mean-pooling is used to fuse the frame embeddings. 
Visual embeddings $\vV_{i}$ can be obtained by:
\begin{equation}\label{eq:meanpooling}
\vV_{i} = \frac{1}{T}\sum_{j}^{T} {\vE_i}(\mathcal{V}_{ij}),
\end{equation}
where $\mathcal{V}_{ij}$ represents the $j_{th}$ frame of $\mathcal{V}_i$. 
The visual to semantic cross-entropy loss $\mathcal{L}_{v2sce}$, $\mathcal{L}_{i2t}$ and $\mathcal{L}_{t2i}$ are adopted to align visual embeddings to semantic features. $\mathcal{L}_{v2sce}$ is similar to $\mathcal{L}_{i2tce}$, defined as:
\begin{equation}\label{eq:lv2sce}
\mathcal{L}_{v2sce}(i) = \sum_{k=1}^N -q_k\log\frac{\exp(s(\vV_i,{\vT}_{y_k}))}{\sum_{{y_j}=1}^N \exp(s(\vV_i,{\vT}_{y_j}))},
\end{equation}
where $q_k$ represents the soft label in the target distribution, and $N$ is the number of identities.
Meanwhile, triplet loss $\mathcal{L}_{tri}$ with soft-margin and ID loss $\mathcal{L}_{id}$ are also used:
\begin{equation}\label{eq:ltri}
\mathcal{L}_{tri} = \max(d_p-d_n+\theta,0),
\end{equation}
\begin{equation}\label{eq:lid}
\mathcal{L}_{id} = \sum_{k=1}^N -q_k\log(p_k),
\end{equation}
where $\theta$ is the soft-margin of $\mathcal{L}_{tri}$, $p_k$ represents ID prediction logits of class $k$,  $d_p$ and $d_n$ are feature distances of positive pair and negative pair. The overall loss $\mathcal L_{stage2}$ is defined as follows:
\begin{equation}\label{eq:ft-stage2loss}
\mathcal L_{stage2} = \mathcal{L}_{v2sce} + \beta \mathcal{L}_{tri} + \gamma \mathcal{L}_{id} + \delta \mathcal{L}_{i2t} + \epsilon \mathcal{L}_{t2i},
\end{equation}
where $\beta$, $\gamma$, $\delta$ and $\epsilon$ balance the importance of the relative losses.

\subsection{Video Set-Level-Adapter for Efficient Model Tuning}
\label{sec: VSLA}
Video ReID requires the model to learn appearance representation in both intra-frame and inter-frames.
We present a novel perspective, where a video sample is regarded as a frame set $\mathcal{S}_i = \{\mathcal{V}_{ij}|j=1,2,...,n\}$ consisting of independent frames, and propose an efficient Video Set-Level-Adapter (VSLA) module. The VSLA consists of two components: an Intra-Frame Adapter (IFA, Fig.~\ref{overall_model} (a)) and a Cross-Frame Attention Adapter (CFAA, Fig.~\ref{overall_model} (b)). IFA is designed to parameter-efficiently adapt the pre-trained visual foundation model to downstream tasks, it takes raw frames as input and provides image-level appearance representation. CFAA takes a set of frames as input, aggregating the inter-frame complementary information for more powerful video-level representations.

IFA consists of two mapping matrices in a bottleneck structure. It runs in parallel with MLP blocks within each layer of the Image Encoder. 
As shown in Fig.~\ref{overall_model}, the Image Encoder in CLIP (ViT-Base-16) consists of alternating layers of Multi-Head Self-Attention (MSA)~\cite{attention_is_all_you}, Multi-Layer Perceptron (MLP) and LayerNorm (LN), which can be formulated as:
\begin{align} \label{ViT_eq}
\mathbf{x}_{i}^{'}  &= \textsc{MSA}(\textsc{LN}(\mathbf{x}_{i-1})) + \mathbf{x}_{i-1}, \\
\mathbf{x}_{i} &= \textsc{MLP}(\textsc{LN}(\mathbf{x}_{i}^{'})) + \mathbf{x}_{i}^{'}.
\end{align}

We denote the input of IFA as $\mathbf{x}_{i}' \in \mathbb{R}^{T \times (N+1) \times D}$, where $N = HW/P^2$, $D$ represents the dimension and $T$ is the number of frames. The down-projection layer $\vW_{down}$ projects $\mathbf{x}_{i}'$ to $\mathbf{x}_{i}'' \in \mathbb{R}^{T \times (N+1) \times \alpha}$, where $\alpha$ is a hyper-parameter. Then $\mathbf{x}_{i}''$ goes through a GELU $\sigma$ and up-projection layer $\vW_{up}$. The process can be formulated as:
\begin{equation}
   \textsc{IFA}(\mathbf{x}_{i}^{'})  = \sigma (\mathbf{x}_{i}^{'} \vW_{down}) \vW_{up}, 
\end{equation}
\begin{equation}
    \mathbf{x}_{i} = \textsc{MLP}(\textsc{LN}(\mathbf{x}_{i}^{'})) + \mathbf{x}_{i}^{'} + \textsc{IFA}(\mathbf{x}_{i}^{'}).
\end{equation}
Unlike LoRA~\cite{LoRA}, which adds trainable pairs of rank decomposition matrices in parallel to every pre-existing weight matrix, IFA is solely in parallel with MLP.
Therefore, adopting IFA results in far fewer parameters, accounting for only 5.5\% ($\alpha=256$) of the whole Image Encoder (ViT-Base-16).


CFAA is also a bottleneck architecture with a cross-frame attention layer in the middle. Our model $\vM(\cdot)$ with CFAA is immune to frame ordering\cite{chao2019gaitset}, which can be formulated as:
\begin{equation}
    \begin{split}
        \vM(\{\mathcal{V}_{ij}|j=1,2,...,n\}) = \vM(\{\mathcal{V}_{i\pi(j)}|j=1,2,...,n\}),
    \end{split}
\end{equation}
where $\pi$ is any permutation\cite{zaheer2017deep}. 
We denote the input of CFAA as $\mathbf{x}_{i-1} \in \mathbb{R}^{T \times (N+1) \times D}$, the down-projection layer projects $\mathbf{x}_{i-1}$ to $\mathbf{x}_{i-1}' \in \mathbb{R}^{T \times (N+1) \times \alpha}$.
The cross-frame attention layer has the same structure as Multi-Head Self-Attention (MSA)\cite{attention_is_all_you}. To aggregate the complementary information among $T$ frames, we reshape the input of cross-frame attention layer $\mathbf{x}_{i-1}'$ to $\mathbf{x}_{i-1}'^{\mathsf{T}}\in \mathbb{R}^{(N+1) \times T \times \alpha}$, and the attention is done in the second dimension of $\mathbf{x}_{i-1}'^{\mathsf{T}}$, thus enabling visual information to exchange across frames. Then, we restore the output of cross-frame attention layer from $\mathbf{x}_{i-1}''^{\mathsf{T}} \in \mathbb{R}^{(N+1) \times T \times \alpha}$ to $\mathbf{x}_{i-1}''\in \mathbb{R}^{T \times (N+1) \times \alpha}$, with $\mathbf{x}_{i-1}''$ passing through up-projection layer. It can be formulated as:
\begin{equation}
    \mathbf{x}_{i}^{'} = \textsc{MSA}(\textsc{LN}(\mathbf{x}_{i-1})) + \mathbf{x}_{i-1} + \textsc{CFAA}(\mathbf{x}_{i-1}).
\end{equation}

\subsection{Platform-Bridge Prompt}
\label{PBP}
We additionally introduce Platform-Bridge Prompt (PBP) to bridge platform differences further. PBP is designed to guide model focusing on platform differences. 
As illustrated in Fig.~\ref{overall_model}, we add a series of platform-specific learnable prompts in the Image Encoder. Specifically, there are only two sets of prompts, one corresponding to the ground platform and the other to the UAV platform.
Applying PBP can be viewed as changing the inputs of each MSA layer in Vision Transformer (ViT~\cite{ViT}). 
We denote the inputs of the MSA layer as $\vh \in \mathbb{R}^{(N+1)\times D}$, where $N = HW/P^2$ and $D$ represents the dimension. The MSA layer with PBP can be formulated as follows,
\begin{align}
f_k(\vh, \vp_k) = \begin{cases}
 MSA_k([\vh:\vp_k^{ground}]) & \text{ if } k<d ~\text{and}~\vh \in Set^{ground} \\
MSA_k([\vh:\vp_k^{uav}]) & \text{ if } k<d ~\text{and}~\vh \in Set^{uav} \\
 MSA_k(\vh) & \text{ if } k\ge d,
\end{cases}
\end{align}
where $\vp_k^{ground} \in\mathbb{R}^{l \times D}, \vp_k^{uav}\in\mathbb{R}^{l \times D}$, $d$ and $l$ are the depth and length of PBP, $[:]$ denotes the concatenation operation, $MSA_k$ represents the $k_{th}$ MSA layer in Image Encoder, $Set^{uav}$ and $Set^{ground}$ are two sets containing the samples from the UAV and the samples from the ground platform respectively. 


\begin{table*}[t]
\centering
\scriptsize
\caption{\footnotesize Comparison with state-of-the-art methods. \dag ~ represents the model initialized by the weight of CLIP\cite{CLIP} released by OpenAI, and \ddag ~ represents the model initialized by weight of ViFi-CLIP\cite{ViFi-CLIP}. We use \textbf{bold} to indicate the best results of our methods, and \underline{underlines} to highlight the best results of other methods. 
On all datasets, our method outperforms the comparisons significantly.} 
\begin{tabular}[t]{C{2.8cm}|C{1.2cm}C{1.2cm}|C{1.2cm}C{1.2cm}|C{1.2cm}|C{1.2cm}C{1.2cm}}
\toprule
\multirow{2}{*}{Method}& \multicolumn{2}{c}{MARS} & \multicolumn{2}{c}{LS-VID} & \multicolumn{1}{c}{iLIDS} & \multicolumn{2}{c}{G2A-VReID}    \\ \cline{2-8}
 & mAP & rank-1 & mAP & rank-1 & rank-1 & mAP & rank-1 \\ \midrule
GLTR\cite{GLTR} & 78.5 & 87.0 & 44.3 & 63.1 & 86 & - & - \\ 
VRSTC\cite{VRSTC} & 82.3 & 88.5 & - & - & 83.4 & - & - \\ 
AP3D\cite{AP3D} & 85.1 & 90.1 & 73.2 & 84.5 & 88.7 & 67.7 & 57.5 \\ 
STGCN\cite{STGCN} & 83.7 & 90.0 & - & - & - & - & - \\ 
MGH\cite{MGH} & 85.8 & 90.0 & - & - & 85.6 & \underline{76.7} & \underline{69.9} \\ 
MG-RAFA\cite{MG-RAFA} & 85.9 & 88.8 & - & - & 88.6 & - & - \\ 
AFA\cite{AFA} & 82.9 & 90.2 & - & - & 88.5 & - & - \\ 
TCLNet\cite{TCLNet} & 85.1 & 89.8 & 70.3 & 81.5 & 86.6 & 65.4 & 54.7 \\ 
STRF\cite{STRF} & 86.1 & 90.3 & - & - & 89.3 & - & - \\ 
GRL\cite{GRL} & 84.8 & 91.0 & - & - & 90.4 & 52.8 & 41.4 \\ 
DenseIL\cite{DenseIL} & \underline{87.0} & 90.8 & - & - & 92 & - & - \\ 
BiCnet-TKS\cite{BiCnet_TKS} & 86.0 & 90.2 & 75.1 & 84.6 & - & 63.4 & 51.7 \\ 
PSTA\cite{PSTA} & 85.8 & 91.5 & - & - & - & 64.6 & 54.5 \\ 
STMN\cite{STMN} & 84.5 & 90.5 & 69.2 & 82.1 & 91.5 & 66.7 & 56.1 \\ 
PiT\cite{PiT} & - & 90.2 & - & - & 92.1 & 76.3 & 67.7 \\ 
SINet\cite{SINet} & 86.2 & 91.0 & 79.6 & 87.4 & \underline{92.5} & 74.5 & 65.6 \\ 
LSTRL\cite{LSTRL} & 86.8 & \underline{91.6} & \underline{82.4} & \underline{89.8} & 92.2 & - & - \\ 
\midrule
FT-CLIP\ddag & 88.00 & 91.62 & 84.07 & 90.77 & 94.00 & 78.11 & 69.32 \\ \midrule
VSLA-CLIP\dag &88.22 & 90.91 & 84.05 & 90.54  & \textbf{95.33} & 79.14 & 71.64 \\ 
VSLA-CLIP\ddag & \textbf{88.60} & \textbf{91.82} & \textbf{85.20} &  \textbf{91.66} & \textbf{95.33}  & \textbf{79.70} & \textbf{72.55} \\ 
 \bottomrule
\end{tabular}
\vspace{-5mm}
\label{tab:compare_with_SOTA}
\end{table*}

\section{Experiments}
In this section, we first introduce the evaluation protocols and implementation details. 
Subsequently, we compare our proposed methods with state-of-the-art algorithms. 
Finally, ablation studies are conducted to investigate the contribution of each component.

\subsection{Datasets and Evaluation Metrics}
We conduct experiments on our G2A-VReID and three widely used video person ReID datasets, $i.e.$, iLIDS\cite{iLIDS}, Mars\cite{mars}, and LS-VID\cite{GLTR_LS-VID}.
For G2A-VReID, we roughly divide 2788 identities into training and test sets at a ratio of $1:2$, similar to that in LS-VID~\cite{GLTR_LS-VID}. 
Therefore, there are 930 identities in training set and 1858 identities in the testing set. 
During the evaluation, we keep the cross-camera search paradigm in ReID task~\cite{GLTR_LS-VID, iLIDS, mars, PRID}. 
Query and gallery are composed of video sequences from the ground and UAV cameras respectively. 

Cumulative Matching Characteristic(CMC) at Rank-1 and mean average precision (mAP) are employed to evaluate the performance of our model.

\subsection{Implementation Details} 
ViT-Base-16~\cite{CLIP} is selected as the Image Encoder. 
The initial weights are chosen as that of ViFi-CLIP~\cite{ViFi-CLIP}, whose Image Encoder and Text Encoder have been fine-tuned on the extensive action recognition dataset Kinetics-400~\cite{kinetics-400}.
Sparse temporal sampling strategy~\cite{wang2016temporal} is used to generate a clip containing 8 frames, with each frame resized to 256$\times$128. We randomly disrupt the order of the frames in each clip. 
Each batch has 32 clips corresponding to 8 identities. Adam~\cite{kingma2014adam} optimizer is used in both stages. 
In the first training stage, we optimize the ID-specific description tokens and shared text prompts with a learning rate of $3.5\times 10^{-4}$, while freezing other parameters. 
In the second training stage, we adopt the initial learning rate $5\times 10^{-6}$ with decaying by 0.1 and 0.01 at the 60$_{th}$ and 90$_{th}$ epoch for FT-CLIP, and the initial learning rate $1 \times 10^{-4}$ with decaying by 0.1 and 0.01 at the 60$_{th}$ and 90$_{th}$ epoch for VSLA-CLIP. 
The margin $\theta$ of triplet loss in Eq.~\eqref{eq:ltri} is set as 0.3, the $\beta$, $\gamma$, $\delta$ and $\epsilon$ in Eq.~\eqref{eq:ft-stage2loss} are 1.0, 0.25, 1.0 and 1.0, respectively. 
Each image is padded with 10 pixels and augmented with random cropping, horizontal flipping, and erasing~\cite{random_ereas}.

\subsection{Comparison with State-of-the-Art Methods}

\textbf{On G2A-VReID Dataset.} 
We comprehensively evaluate nine state-of-the-art methods~\cite{SINet, PiT, STMN, PSTA, BiCnet_TKS, GRL, AP3D, MGH, TCLNet} on G2A-VReID, and report the results in Tab.~\ref{tab:compare_with_SOTA}. 
As can be seen that, MGH\cite{MGH} and PiT~\cite{PiT} showed superior performances on our G2A-VReID dataset, \ie MGH achieves 76.7\% on mAP and 69.9\% on Rank-1. 
We attribute this to the careful visual alignment strategy adopted by MGH and PiT, which involves splitting the full image into vertical or horizontal stripes and aiming to align the stripes.
This strategy mitigates the challenges of self-occlusion inherent in the UAV perspective.
Our method, extracting description tokens for each person and aligning visual embeddings with semantic features, effectively solves the cross-platform visual misalignment problem. VSLA-CLIP\ddag~ achieves 79.70\% mAP and 72.55\% Rank-1 on G2A-VReID.

\textbf{On All Video ReID Dataset.} 
As shown in Tab.~\ref{tab:compare_with_SOTA}, all the variants of our methods with aligning visual embeddings to semantic features, show consistent improvement on all datasets. 
Especially, our method achieves 85.20\% mAP and 91.66\% Rank-1 on the challenging LS-VID dataset, which greatly improves the mAP by 2.80\% and the Rank-1 by 1.86\% compared with the state-of-the-art LSTRL~\cite{LSTRL}. 
\textbf{2)} Models initialized by weights of ViFi-CLIP (ViFi-weight) are marked as $\ddag$, and it is effective compared with the original model weights released by Open AI (marked as $\dag$). 
\textbf{3)} It is worth noting that VSLA-CLIP shows better performance than fine-tuning the whole Image Encoder (FT-CLIP),  with far fewer tunable parameters.
Specifically, VSLA-CLIP$\ddag$ outperforms the FT-CLIP$\ddag$ by 1.59\% mAP on G2A-VReID with tuning parameters (14.5M vs 88.0M). 

Our experiments show that adapting pre-trained image-based models to video ReID tasks with the Video Set-Level-Adapter is both effective and efficient, setting a new baseline method for research endeavors in this field.

\begin{table*}[t]
\centering
\scriptsize
\caption{\footnotesize Effectiveness of proposed components and comparison of the number of tunable parameters. \textbf{baseline} represents training FT-CLIP\ddag~ without Lv2sce in Eq.\eqref{eq:lv2sce}, \textbf{VSA} is Visual-Semantic Alignment, IFA represents Intra-Frame Adapter, CFAA is Cross-Frame Attention Adapter and PBP is Platform Bridge Prompt.}
\begin{tabular}[t]{l|C{1.5cm}C{1.5cm}|C{1.0cm}C{1.0cm}|C{0.9cm}C{0.9cm}C{0.9cm}C{0.9cm}}
\toprule
\multirow{2}{*}{Methods} & Overall & Tunable & \multicolumn{2}{c|}{LS-VID} & \multicolumn{2}{c}{G2A-VReID} \\ \cline{4-7}
& Param(M)& Param (M) & mAP & rank-1 & mAP & rank-1 \\ 
\midrule
AP3D\cite{AP3D} & 34.0 & 24.9 & 73.2 & 84.5 & 67.7 & 57.5 \\
BiCnet-TKS\cite{BiCnet_TKS} & 33.7 & 29.3 & 75.1 & 84.6 & 63.4 & 51.7\\
STMN\cite{STMN} & 90.9 & 87.0 & 69.2 & 82.1 & 66.7 & 56.1\\
SINet\cite{SINet} & 33.7 & 27.3 & 79.6 & 87.1 & 74.5 & 65.6\\ \midrule
baseline & 86.1 & 86.1 & 76.10 & 84.26 & 72.80 & 63.62 \\
baseline+VSA (FT-CLIP\ddag) & 127.4 & 88.0 & 84.07 & 90.77 & 78.11 & 69.32\\ \midrule
IFA & 90.8 & \textbf{4.7} & 77.31 & 84.86 & 73.82 & 65.12  \\
IFA+VSA & 132.1 & 6.6 & 84.16 & 90.94 & 79.01 & 71.67  \\
IFA+VSA+CFAA (VSLA-CLIP\ddag) & 140.0 & 14.5 & \textbf{85.20} & \textbf{91.66} & 79.70 & 72.55 \\ 
IFA+VSA+CFAA+PBP & 140.0 & 14.5 & - & - & \textbf{81.29} & \textbf{74.27} \\ \bottomrule
\end{tabular}
\label{tab:ablation}
\end{table*}

\subsection{Ablation Study}
To demonstrate the effectiveness of our proposed components in Sec.\ref{sec:approch}, we conduct ablation studies and compare our method with four other methods. 

\textbf{Effectiveness of Visual-Semantic Alignment.} To verify the effectiveness of Visual-Semantic Alignment, we first fine-tune the Image Encoder by directly using two common losses ($\mathcal{L}_{tri}$ and $\mathcal{L}_{id}$ in Eq.\eqref{eq:ft-stage2loss}), and set this model as our baseline. As shown in Tab.~\ref{tab:ablation}, Visual-Semantic Alignment is effective for both finetuning-based methods (FT-CLIP\ddag~ vs. baseline) and adapter-based methods (IFA+VSA vs. IFA). In addition, we further analyze the loss functions for visual-semantic alignment. As shown in Tab.~\ref{tab: loss}, when $\mathcal{L}_{i2t}$, $\mathcal{L}_{t2i}$ and $\mathcal{L}_{v2sce}$ are used jointly, our model achieves the best results on LS-VID.
\begin{table}[t]
    \begin{minipage}[t]{0.58\linewidth}
    \caption{Effect of $\alpha$ of Intra-Frame Adapter and Cross-Frame Attention Adapter on LS-VID. TP represents the tunable parameter.}
    \label{tab: alpha}
    \resizebox{0.98\linewidth}{!}{
        \tiny
        \begin{tabular}{C{0.6cm}|C{0.6cm}|C{0.6cm}|C{0.8cm}C{0.8cm}}
            \hline
            \multirow{2}{*}{$\alpha$} & \multicolumn{2}{c|}{TP (M)} & \multicolumn{2}{c}{LS-VID} \\ \cline{2-5}
                   & IFA               & CFAA               & mAP        & rank-1        \\
            \hline
            64  & 1.2 & 1.4 & 79.58 & 86.71\\
            128 & 2.4 & 3.2 & 83.64 & 90.00\\
            256 & 4.7 & 7.9 & \textbf{85.20} & \textbf{91.66} \\
            384 & 7.1 & 14.2 & 85.09 & 91.49 \\
            \hline
        \end{tabular}
    }
    \end{minipage}
    \begin{minipage}[t]{0.41\linewidth}
    \caption{Ablation experiments for the losses used for Visual-Semantic Alignment on LS-VID.}
    \label{tab: loss}
    \resizebox{0.92\linewidth}{!}{
        \scriptsize
        \begin{tabular}{C{1.0cm}|C{0.6cm}|C{0.6cm}|C{0.8cm}}
            \hline
            $\mathcal{L}_{v2sce}$ & $\mathcal{L}_{i2t}$ & $\mathcal{L}_{t2i}$ & mAP\\
            \hline
             \checkmark &  &  & 84.73\\
             & \checkmark  & & 84.29\\
             &  & \checkmark & 84.45 \\
             &\checkmark  & \checkmark & 84.71\\
            \checkmark  & \checkmark  &\checkmark  & \textbf{85.20}\\
            \hline
        \end{tabular}
    }
    \end{minipage}
\end{table}

\textbf{Effectiveness of Video Set-Level-Adapter.}
Our goal for proposing the Video Set-Level-Adapter is to efficiently adapt pre-trained image-based visual foundation mode to video-based ReID tasks. 
Considering that the Video Set-Level-Adapter (VSLA) contains two modules, $i.e.$, an Intra-Frame Adapter (IFA) and a Cross-Frame Attention Adapter (CFAA), we perform ablation experiments separately to verify the effectiveness of each module.
As shown in Tab.~\ref{tab:ablation}, IFA surpasses the full fine-tuned baseline method (77.31\% vs. 76.10\% mAP on LS-VID) with significantly less number of tunable parameters (4.7M vs. 86.1M). 
In addition, CFAA further improves model performance (85.20\% vs. 84.16\% mAP on LS-VID) while also using a small number of tunable parameters, which indicates that regarding video sequences as a set is effective.


\begin{figure*}[h]
	\centering
	\includegraphics[width=0.5\textwidth]{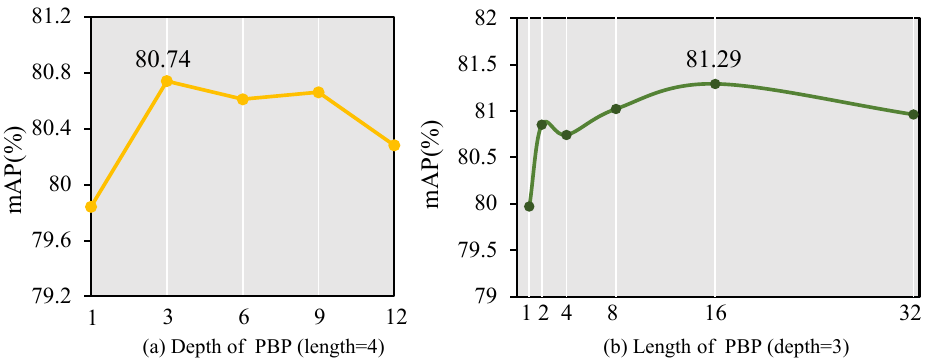}
	\caption{Analysis on the depth and length of PBP on our G2A-VReID.}
	\label{fig:abalation_PBP}
\end{figure*}
We also analyze the hyper-parameter $\alpha$ introduced in Sec.~\ref{sec: VSLA}, which determines model's complexity and the number of training parameters.
We set $\alpha$ to be 64, 128, 256, and 384 respectively. 
As presented in Tab.~\ref{tab: alpha}, the performances tend to improve with increasing $\alpha$, and achieves the best mAP at $\alpha = 256$. Therefore, we fix $\alpha$ to be 256 for other datasets. At this setting, the VSLA module contains only approximately 12.6 million parameters parameters, and VSLA-CLIP achieves 85.20\% mAP on LS-VID, surpassing FT-CLIP by 1.13\%.

\textbf{Effectiveness of PBP.} 
The Platform Bridge Prompt (PBP) offers meticulous instructions to enable models to discern differences across platforms. 
It adeptly steers the model towards obtaining precise and targeted information, thereby bridging the semantic gap in visual features.
The depth $d$ and length $l$ are two hyper-parameters in PBP, which are introduced in Sec.~\ref{PBP}. To analyze the impact of these two parameters on the model, we use grid-search to explore the impact of different value combinations on the model performance. 
The results for various parameter combinations of the model are presented in Fig.~\ref{fig:abalation_PBP}, and the optimal performance is achieved when $d=3$ and $l=16$.

\section{Conclusion}
In this paper, we construct a large-scale benchmark dataset for cross-platform video person ReID. 
Besides, we also propose a baseline method for solving cross-platform visual misalignment problems by transforming the visual alignment problem into visual-semantic alignment through the vision-language model ($i.e.$, CLIP).
To efficiently and effectively adapt the pre-trained image-based visual foundation model to Video ReID, We propose a Video Set-Level-Adapter module, which aggregates the inter-frame complementary information for more powerful video-level representations with only 12.6 million trainable parameters. 
Experimental results demonstrate that our proposed methods achieve state-of-the-art performance and will be a new trend for cross-platform video ReID tasks. 
\subsection*{Acknowledgement}
This work was supported in part by the National Natural Science Foundation of China (NSFC) under Grant 62101453, 62176198 and 62201467, the Key Research and Development Program of Shaanxi Province under Grant 2024GX-YBXM-135, in part by China Postdoctoral Science Foundation under Grant 2022TQ0260, 2023M742842, in part by the Young Talent Fund of Xi'an Association for Science and Technology under Grant 959202313088, Innovation Capability Support Program of Shaanxi (No. 2024ZC-KJXX-043), the Fundamental Research Funds for the Central Universities No. HYGJZN202331 and the Natural Science Basic Research Program of Shaanxi Province (No. 2022JC-DW-08).

%
%
\bibliographystyle{splncs04}

\end{document}